\documentclass[letterpaper]{article} % DO NOT CHANGE THIS
\usepackage{aaai2027}  % arXiv non-anonymous preprint
\nocopyright
\usepackage[hyphens]{url}  % DO NOT CHANGE THIS
\usepackage{graphicx} % DO NOT CHANGE THIS
\usepackage{natbib}  % DO NOT CHANGE THIS AND DO NOT ADD ANY OPTIONS TO IT
\usepackage{caption} % DO NOT CHANGE THIS AND DO NOT ADD ANY OPTIONS TO IT
\usepackage{algorithm}
\usepackage{algorithmic}

\usepackage{newfloat}
\usepackage{listings}
\DeclareCaptionStyle{ruled}{labelfont=normalfont,labelsep=colon,strut=off} % DO NOT CHANGE THIS
\floatstyle{ruled}
\newfloat{listing}{tb}{lst}{}
\floatname{listing}{Listing}

\usepackage{booktabs}

\usepackage{amsmath}
\usepackage{amssymb}
\usepackage{multirow}
\usepackage{xcolor}
\usepackage{pifont}
\title{FinProBench: Evaluating Financial AI Agents with Role-Grounded Rubrics Derived from Professional Deliverables}

\author{
Ben Wang\textsuperscript{\rm 1*},
Kang Zhou\textsuperscript{\rm 1},
Lifan Guo\textsuperscript{\rm 1},
Feng Chen\textsuperscript{\rm 1},
Chi Zhang\textsuperscript{\rm 1}
}
\affiliations{
\textsuperscript{\rm 1}Qwen DianJin Team, Alibaba Cloud Computing\\
wangben1619@gmail.com,\\
\{wuyue.zk, lifan.lg, betterman.chenf, edward.zhang\}@alibaba-inc.com
}

\begin{document}

\maketitle

\begin{abstract}
Evaluating AI agents on professional financial tasks requires assessment criteria that reflect the quality standards of real work products. Most rubric construction methods, whether direct generation, contrastive extraction, or iterative refinement, build criteria from task prompts or model outputs. As a result, they miss the tacit professional standards that appear only in real deliverables. We introduce \textbf{FinProBench}, a benchmark for evaluating AI agents on professional financial job tasks, together with a \textbf{Role-Grounded Rubric Construction} (RGRC) pipeline designed for reuse across roles, tasks, and jurisdictions. Rather than relying on prompts or model outputs, RGRC distills evaluation criteria from real deliverables written by practitioners in the same occupational role. The pipeline runs in four stages: \textit{Deliverable Collection}, \textit{Competency Extraction}, \textit{Rubric Synthesis}, and \textit{Validation}. The resulting rubrics are comprehensive, discriminative, and transferable: they capture tacit professional standards, separate quality levels, and apply across tasks within the same role. Before analysis, we classified all 57 occupations by deliverable-genre characteristics into 30 prior-rich conventional roles and 27 prior-sparse role-specialized roles. Across all 30 conventional and 27 role-specialized roles, Prompt-only nearly matches RGRC in the conventional group (89.2\% versus 90.7\%), whereas RGRC reaches 99.1\% versus 78.0\% in the role-specialized group. This regime split shows that prompt engineering can approximate rubrics when conventions are already encoded in general-model priors, whereas authentic professional grounding is critical for recovering tacit, role-specific standards beyond those priors. FinProBench draws on an evidence corpus of 1,723 curated real-world deliverables spanning 57 financial occupations, 8 sub-industries, and 161 deliverable types, from which we release an initial evaluation set of 20 complete tasks covering 20 roles across 7 sub-industries. Using a panel of heterogeneous LLM judges under an unbiased role-level rubric, authentic human deliverables rank first on average (73.7 versus\ 70.3, 70.2, and 69.6 on a 100-point scale), and the four systems exhibit overlapping 95\% confidence intervals and complementary dimension-level strengths. Because rubrics are reused at the role level, per-task construction effort drops by an estimated 6.7$\times$ compared with authoring each rubric from scratch.
\end{abstract}

% Uncomment the following to link to your code, datasets, an extended version or similar.
% You must keep this block between (not within) the abstract and the main body of the paper.
% Make sure that you do not de-anonymize yourself with these links.
% \begin{links}
%     \link{Code}{https://aaai.org/example/code}
%     \link{Datasets}{https://aaai.org/example/datasets}
%     \link{Extended version}{https://aaai.org/example/extended-version}
% \end{links}
\section{Introduction}

The rapid advancement of large language models (LLMs) \citep{openai2023gpt4} has catalyzed their deployment as AI agents performing complex professional tasks \citep{ouyang2022training, bai2022constitutional}. Recent evaluation efforts on general-purpose knowledge and reasoning \citep{hendrycks2021mmlu, srivastava2022bigbench, rein2023gpqa, wang2024mmlupro} and on holistic model behavior \citep{liang2023helm} have demonstrated impressive capabilities of frontier models. In the financial industry, where precision, regulatory compliance, and domain expertise are paramount, evaluating whether AI agents can produce work at professional standards presents unique challenges that existing benchmarks inadequately address.

\paragraph{Limitations of Current Financial AI Benchmarks.}
Financial NLP benchmarks have evolved from sentiment classification and NER \citep{shah2022flue, lei2023cfbenchmark} to reasoning and QA \citep{islam2023financebench, xie2024finben, xie2023pixiu}. Domain-specific benchmarks such as FinEval \citep{guo2023fineval} and CFinBench \citep{nie2024cfinbench} evaluate financial knowledge through examinations, while models like BloombergGPT \citep{wu2023bloomberggpt}, FinGPT \citep{yang2023fingpt}, and DISC-FinLLM \citep{chen2023discfinllm} perform well on such tasks. Yet a broader methodological gap across markets persists: \textit{existing benchmarks rarely evaluate AI agents on producing complete professional deliverables}, the multi-page research reports, valuation models, and compliance analyses that constitute actual work output. FinBen \citep{xie2024finben}, while comprehensive with 35 datasets, remains confined to extractive evaluations. FinanceBench \citep{islam2023financebench} grounds questions in real filings but limits responses to short factual answers. This mirrors limitations in law \citep{guha2023legalbench}, medicine \citep{jin2020medqa}, and science \citep{wang2024scibench}, where benchmarks test factual recall rather than professional artifact production.

\paragraph{The Evaluation Crisis: From Answers to Deliverables.}
The emergence of AI agents producing long-form, structured work products \citep{xu2024theagentcompany, li2026jobbench, liu2023agentbench} has created an urgent need for evaluation beyond binary correctness. Benchmarks like SWE-bench \citep{jimenez2024swebench}, GAIA \citep{mialon2024gaia}, WebArena \citep{zhou2024webarena}, and OSWorld \citep{xie2024osworld} have pushed the frontier of consequential agent evaluation. GDPval \citep{patwardhan2025gdpval} pioneered economically grounded evaluation on 220 real-world tasks across 9 sectors. Its valuable finance tasks use separately authored expert rubrics, however, leaving no systematic mechanism for evidence-grounded derivation or role-level reuse across tasks, occupations, and jurisdictions. More recent work including TheAgentCompany \citep{xu2024theagentcompany}, Agents' Last Exam \citep{sun2026ale}, and JobBench \citep{li2026jobbench} has demonstrated feasibility of evaluating agents on workplace tasks but relies primarily on programmatic verification rather than rubric-based quality assessment of open-ended deliverables.

\paragraph{The Rubric Construction Challenge.}
Rubric-based evaluation has emerged as the dominant paradigm for assessing open-ended LLM outputs \citep{zheng2023judging, lin2025wildbench, liu2023geval, kim2024prometheus}, complementing reward-model scoring \citep{lambert2024rewardbench}, preference optimization \citep{rafailov2023dpo}, and crowdsourced comparison \citep{chiang2024chatbotarena, bradley1952rank}. Yet constructing high-quality rubrics remains a critical bottleneck \citep{liu2026rubricsurvey}. \textit{Direct generation} produces criteria in a single LLM pass but yields superficial, prompt-parroting results. \textit{Contrastive generation} \citep{liu2025openrubrics} extracts criteria from preference pairs, but salient differences between responses may be stylistic rather than substantive \citep{shen2026rrd}. \textit{Iterative refinement}, including Auto-Rubric \citep{xie2025autorubric}, OptimSyn \citep{fan2026optimsyn}, and RRD \citep{shen2026rrd}, represents significant advances but still derives rubrics from model outputs. Online methods like SibylSense \citep{xu2026sibylsense} and ARES \citep{li2026ares} share the same limitation: \textbf{to the best of our knowledge, these methods construct rubrics from task prompts or model-generated responses, rather than from authentic professional work products.} Concurrent work in financial deep research, most notably FinResearchBench~II \citep{luan2026finresearchbench2}, likewise removes human experts from the rubric loop, but derives criteria \emph{from model-generated reports}. We instead ground rubrics in \emph{authentic human deliverables}, which additionally furnish an authentic human-deliverable anchor for validation that report-derived benchmarks lack.

\paragraph{From Flat Checklists to Two-Level Structures.}
Parallel to grounding, recent work highlights \emph{rubric structure} as a distinct axis of quality. RubricHub \citep{liauto2026rubrichub} introduces a coarse-to-fine framework that first elicits broad quality dimensions and then discovers fine-grained items able to separate near-tie responses, curating $\sim$110K rubrics for reward-model training. RubricBench \citep{rubricbench2026} empirically measures a 27\% gap between LLM-generated and expert-authored rubrics, attributing the failures to \emph{attention drift} and \emph{value inversion}, symptoms of flat, weight-uniform checklists. In the same spirit, DRACO \citep{perplexity2026draco} scores long-form deep-research reports along four weighted meta-dimensions (accuracy, completeness, objectivity, citation quality), evidencing that per-dimension aggregation is standard practice for long-form evaluation. RGRC complements these efforts but differs in its source of evidence. Existing methods add structure to model-generated content. RGRC instead grounds both the meta-dimensional priors and the specific criteria in authentic professional deliverables. It also extends the two-level design from a single evaluation protocol to a role-grounded rubric structure reusable across all tasks of an occupation.

\begin{figure}[t]
\centering
\includegraphics[width=0.86\columnwidth]{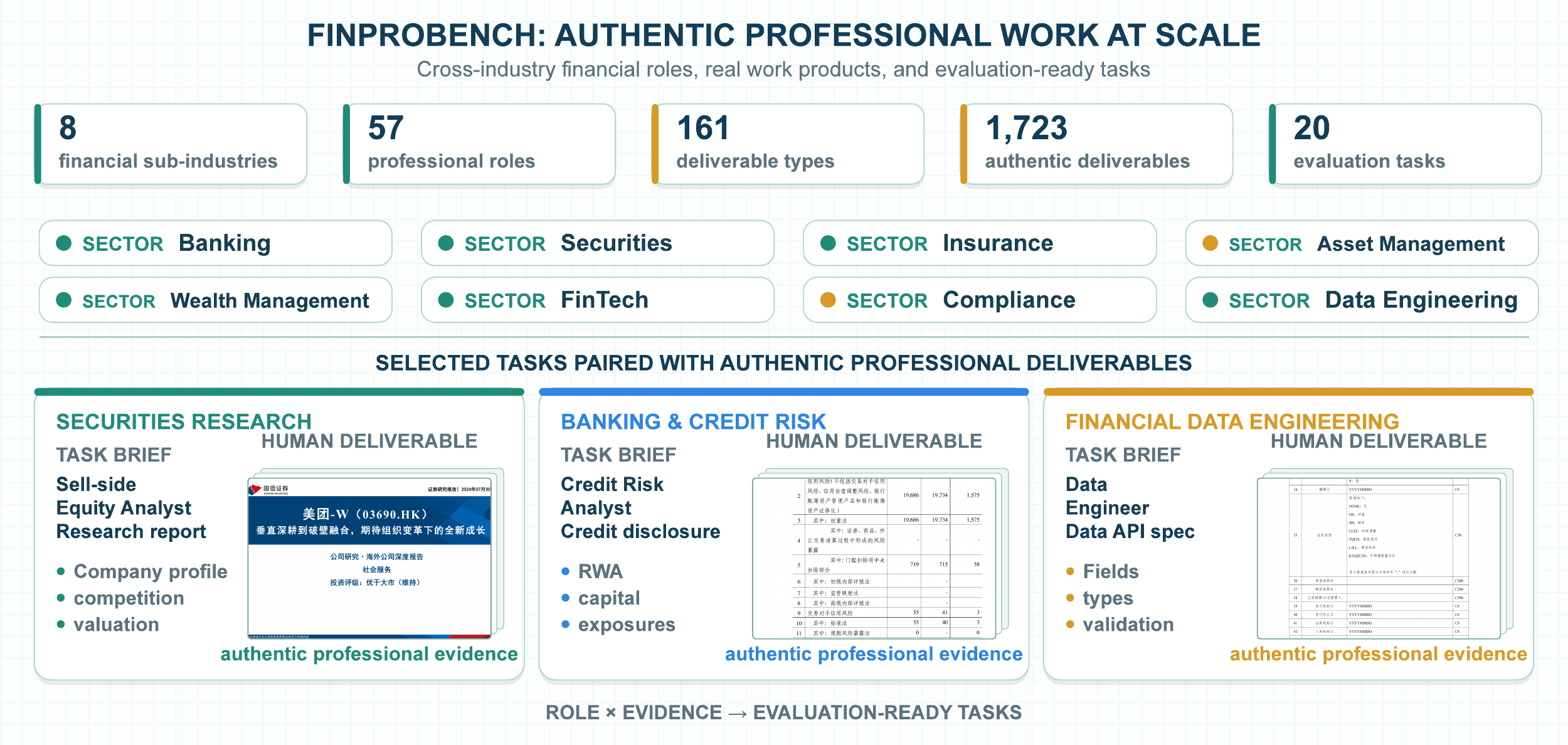}
\caption{FinProBench coverage: 1,723 professional deliverables spanning 57 roles, 161 deliverable types, and 8 financial sub-industries; the initial evaluation set contains 20 tasks across 20 roles and 7 sub-industries.}
\label{fig:coverage}
\end{figure}

\paragraph{Our Contribution.}
We present \textbf{FinProBench}, a benchmark and methodology that simultaneously addresses all three gaps through \textbf{Role-Grounded Rubric Construction} (RGRC). Our key contributions are:

\begin{enumerate}
    \item \textbf{A novel rubric construction paradigm} grounded in authentic professional deliverables. All 57 occupations were classified before analysis from deliverable-genre characteristics into 30 prior-rich conventional roles and 27 prior-sparse role-specialized roles. Across all 30 conventional and 27 role-specialized roles, Prompt-only nearly matches RGRC in the conventional group (89.2\% versus 90.7\%), whereas RGRC leads by 21.2 points in the role-specialized group (99.1\% versus 78.0\%).
    \item \textbf{A four-stage, predominantly automated pipeline} (\textit{Deliverable Collection} $\rightarrow$ \textit{Competency Extraction} $\rightarrow$ \textit{Rubric Synthesis} $\rightarrow$ \textit{Validation}) that produces \textbf{hierarchical rubrics}, a \emph{role-level normative layer} shared across all tasks of an occupation plus a thin \emph{task-specific layer}, both generated by the same pipeline. This layering delivers full prompt--rubric coverage and an estimated $6.7\times$ reduction in per-task authoring effort through role-level reuse.
    \item \textbf{A reusable financial professional-deliverable benchmark} built from an evidence corpus of 1,723 curated real-world deliverables spanning 57 occupations, 8 sub-industries (banking, securities, insurance, asset management, wealth management, fintech, compliance, and data engineering), and 161 deliverable types, from which we release an initial evaluation set of 20 complete tasks covering 20 roles across 7 sub-industries (Figure~\ref{fig:coverage}). This instantiation demonstrates how RGRC turns a role-structured evidence corpus into a scalable evaluation suite rather than relying on independently hand-authored rubrics for every task.
    \item \textbf{A scalable validation protocol} combining a four-family judge panel, gold-held-out human anchors, and self-preference checks. The resulting evaluation exhibits substantial cross-evaluator agreement ($\kappa=0.76$) and stable system rankings under judge-panel ablations.
\end{enumerate}

\section{Method: Role-Grounded Rubric Construction}

Our method is premised on a key insight: professionals within the same occupational role share implicit quality standards that are \textit{embodied in their work products} but rarely articulated in task descriptions. A senior equity research analyst's report exhibits consistent structural patterns, analytical depth, and evidentiary standards regardless of the specific company under coverage. These embodied standards constitute the professional-quality reference against which AI agent outputs should be evaluated.

\subsection{Overview}

\begin{figure*}[t]
\centering
\includegraphics[width=0.94\textwidth]{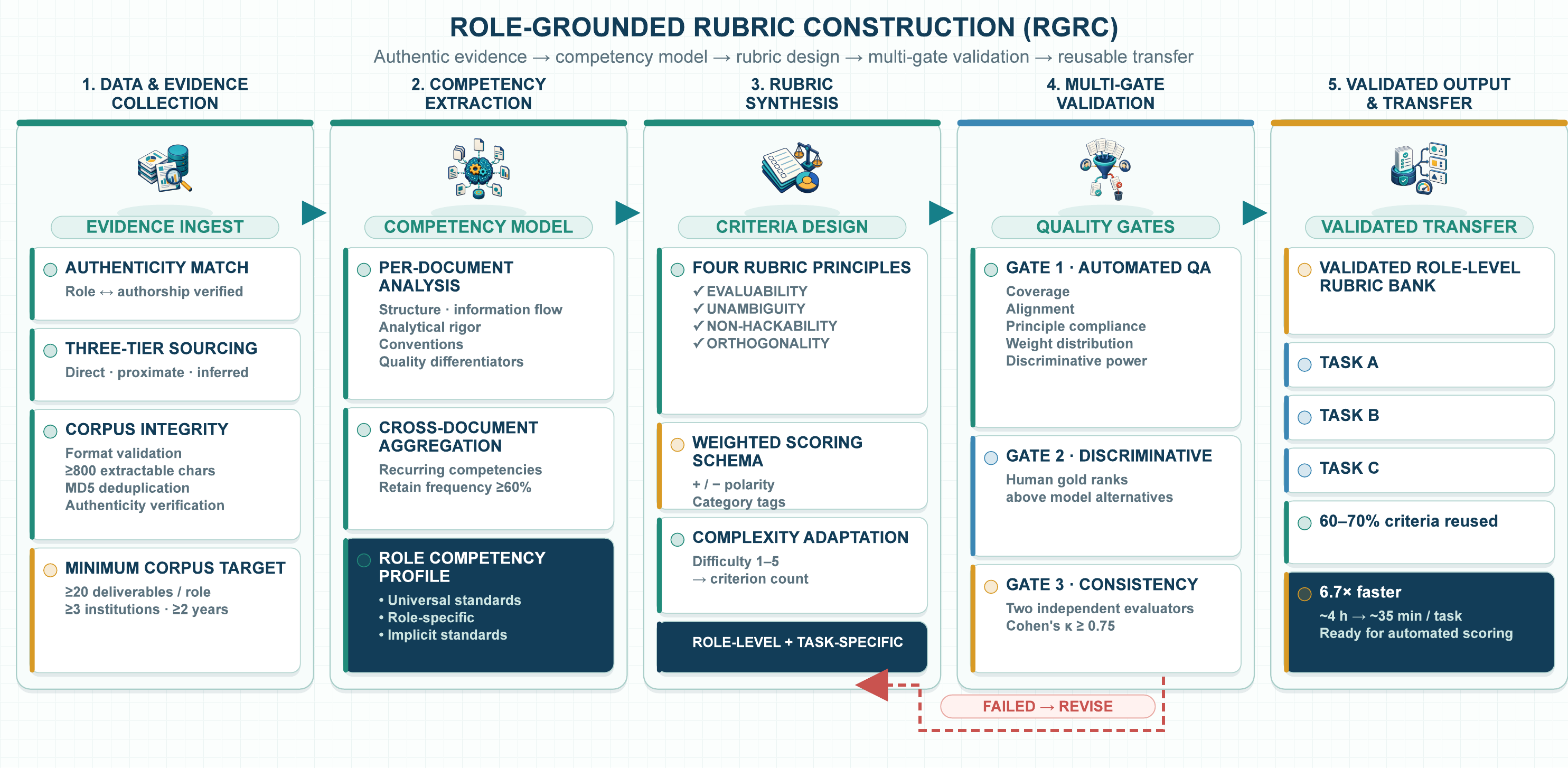}
\caption{The Role-Grounded Rubric Construction (RGRC) pipeline. Starting from a
corpus of \emph{authentic} professional deliverables produced by practitioners
in the same occupational role, the pipeline proceeds through four stages.
\textbf{Stage~1 (Deliverable Collection)} gathers genuine work products under an
authenticity criterion and a three-tier source-prioritization scheme, filtering
out prompts, textbook exemplars, and synthetic artifacts.
\textbf{Stage~2 (Competency Extraction)} first performs per-document analysis to
surface the tacit skills, evidentiary standards, and structural conventions each
deliverable embodies, then cross-document synthesis to consolidate them into a
role-level competency profile.
\textbf{Stage~3 (Rubric Synthesis)} converts this profile into candidate criteria
governed by four design principles, \emph{evaluability}, \emph{unambiguity},
\emph{non-hackability}, and \emph{orthogonality}, together with a scoring schema
and a complexity-adaptive criterion count.
\textbf{Stage~4 (Validation)} admits rubrics through three sequential
gates: automated quality assurance, discriminative review, and a
cross-evaluator consistency gate ($\kappa \ge 0.75$); rubrics failing the gates are
returned to Stage~3 through a bounded synthesis-retry loop (dashed path).
Validated role-level rubrics are finally specialized into task-level rubrics,
inheriting 60--70\% of criteria and reducing end-to-end construction time from $\sim$4 hours to $\sim$35 minutes (an estimated 6.7$\times$) per task.}
\label{fig:pipeline}
\end{figure*}

The Role-Grounded Rubric Construction (RGRC) pipeline consists of four stages, illustrated in Figure~\ref{fig:pipeline}:

\begin{enumerate}
    \item \textbf{Stage 1: Deliverable Collection.} Curate a corpus of $\geq 20$ authentic professional deliverables per occupational role from authoritative public sources.
    \item \textbf{Stage 2: Competency Extraction.} Extract latent quality dimensions and professional standards from the deliverable corpus using LLM-based analysis grounded in the actual work products.
    \item \textbf{Stage 3: Rubric Synthesis.} Synthesize extracted competencies into structured, scored rubric criteria following four design principles.
    \item \textbf{Stage 4: Validation.} Subject generated rubrics to automated quality assurance and discriminative review with iterative refinement until convergence.
\end{enumerate}

%\begin{figure}[t]
%\centering
%\fbox{\parbox{0.95\columnwidth}{\centering
%\small
%\textbf{Stage 1: Deliverable Collection}\\
%$\downarrow$ \textit{20+ authentic deliverables per role}\\
%\textbf{Stage 2: Competency Extraction}\\
%$\downarrow$ \textit{latent quality dimensions}\\
%\textbf{Stage 3: Rubric Synthesis}\\
%$\downarrow$ \textit{scored evaluable criteria}\\
%\textbf{Stage 4: Expert Validation}\\
%$\downarrow$ \textit{expert-approved rubrics}\\
%\textbf{Evaluation Task Package}
%}}
%\caption{The four-stage RGRC pipeline. Each stage is designed to be automated with LLM assistance while maintaining expert oversight at critical decision points.}
%\label{fig:pipeline}
%\end{figure}

\subsection{Stage 1: Deliverable Collection}
\label{sec:stage1}

The foundation of RGRC is a carefully curated corpus of \textit{authentic professional deliverables}, documents that were actually produced by practitioners in the target occupational role as part of their professional work. This distinguishes our approach from methods that use synthetic examples or model-generated references.

\paragraph{Authenticity Criterion.}
We define a document as an authentic deliverable for role $r$ if and only if:
\begin{equation}
    \text{role}(\text{author}(d)) \approx r
\end{equation}
That is, the person who authored document $d$ occupies a role approximately equivalent to the target role $r$. This criterion explicitly excludes:
\begin{itemize}
    \item Academic papers (authored by researchers, not practitioners)
    \item Regulatory documents (authored by regulators, not market participants)
    \item Annual reports (authored by IR departments, not analysts)
    \item Industry association publications (authored by associations, not practitioners)
\end{itemize}

\paragraph{Source Prioritization.}
We establish a three-tier source hierarchy based on authenticity confidence:
\begin{enumerate}
    \item \textbf{Tier 1 (Direct):} Documents with clear institutional authorship matching the role (e.g., sell-side research reports from securities firms for equity analyst roles; API documentation from exchange technology departments for data engineering roles).
    \item \textbf{Tier 2 (Proximate):} Documents from adjacent roles with substantial overlap (e.g., consulting firm solution whitepapers for solution architect roles).
    \item \textbf{Tier 3 (Inferred):} Documents where authorship role is inferred from content characteristics and publication context.
\end{enumerate}

\paragraph{Corpus Requirements.}
For each role $r$, we require a minimum of 20 deliverables from at least 3 distinct institutional sources, spanning $\geq 2$ years to capture evolving standards, with format diversity (e.g., both periodic reports and event-driven analyses for research roles).

In practice, we curated 1,723 deliverables across 57 financial occupations and 161 deliverable types, with a minimum of 20 deliverables per occupational role in the core coverage matrix after quality filtering; the per-role corpus is pooled across the role's deliverable types rather than requiring 20 documents for every individual type (see the Benchmark section for details).

\subsection{Stage 2: Competency Extraction}
\label{sec:stage2}

Given a deliverable corpus $\mathcal{D}_r = \{d_1, d_2, \ldots, d_n\}$ for role $r$, we extract latent quality dimensions through a two-phase process.

\paragraph{Phase 2a: Per-Document Analysis.}
For each deliverable $d_i$, we prompt an LLM to identify structural patterns (document organization, information flow), analytical depth markers (multi-step reasoning, quantitative rigor), professional conventions (domain terminology, formatting standards), and quality differentiators (elements distinguishing professional-grade work from superficial analysis). Formally:
\begin{equation}
    C_i = \text{LLM}_{\text{extract}}(d_i, \text{role\_context}(r))
\end{equation}

\paragraph{Phase 2b: Cross-Document Synthesis.}
We aggregate per-document competencies to identify \textit{role-level} standards, quality dimensions that appear consistently across deliverables from multiple authors and institutions:
\begin{equation}
    C_r^{\text{role}} = \text{Aggregate}(\{C_1, C_2, \ldots, C_n\}, \tau_{\text{freq}})
\end{equation}
where $\tau_{\text{freq}}$ is a frequency threshold ensuring that retained competencies appear in $\geq 60\%$ of analyzed deliverables. This filtering eliminates idiosyncratic author preferences while preserving genuine professional standards.

The synthesis distinguishes universal competencies (shared across all financial roles, e.g., numerical accuracy), role-specific competencies (e.g., DCF model coherence for valuation analysts), and implicit standards, requirements never stated in prompts but consistently demonstrated in expert work (e.g., sell-side reports invariably include risk disclosure sections).

\subsection{Stage 3: Rubric Synthesis}
\label{sec:stage3}

Extracted competencies are transformed into scored rubric criteria following four design principles inspired by \citet{shen2026rrd}, \citet{li2026ares}, and the broader alignment literature \citep{ouyang2022training, bai2022constitutional, rafailov2023dpo}:

\paragraph{Principle 1: Evaluability.}
Each criterion is an independently judgeable unit with a determinate scoring rule, a clear boundary and an objective decision procedure. Graded criteria are acceptable (e.g., ``lists volatility, drawdown, Sharpe ratio, and Beta,'' scored 0--4 by the number covered); what we exclude are anchorless subjective descriptions (e.g., ``the analysis is deep and thorough'') that provide no verifiable basis for scoring. We deliberately avoid forcing decomposition to single facts, which in complex financial tasks explodes into unwieldy rubrics of 1000+ items.

\paragraph{Principle 2: Unambiguity.}
Criteria are formulated such that two independent evaluators would assign the same score with probability $\geq 0.9$. We achieve this through:
\begin{itemize}
    \item Concrete observable indicators rather than subjective quality terms
    \item Quantitative thresholds where applicable (e.g., ``Brinson attribution components sum to within 0.3 percentage points of total excess return'')
    \item Binary or low-cardinality scoring (predominantly $[+1]$ and $[+2]$ weights)
\end{itemize}

\paragraph{Principle 3: Non-Hackability.}
Criteria resist ``rubric gaming'', strategies that satisfy literal criteria without demonstrating genuine competence. Following \citet{li2026jobbench}'s chained rubric design, we include criteria that require multi-step reasoning chains where intermediate results constrain downstream answers:
\begin{equation}
    \text{score}(c_j) > 0 \implies \text{consistent}(c_j, \{c_k : k \in \text{deps}(j)\})
\end{equation}

\paragraph{Principle 4: Orthogonality.}
Criteria should be minimally correlated. Scoring high on one criterion should not mechanically guarantee scoring high on another. We verify orthogonality through correlation analysis on pilot scoring runs.

\paragraph{Scoring Schema.}
Each criterion $c_j$ is assigned a weight $w_j \in \{1, 2, 3, 5\}$ (minor detail to defining quality), a polarity (positive $[+w_j]$ for desired properties, negative $[-w_j]$ for catastrophic failures), and optional category tags (\texttt{factual}, \texttt{analytical}, \texttt{structural}, \texttt{compliance}, \texttt{format}).

The total rubric score for a task is:
\begin{equation}
    S_{\text{max}} = \sum_{j: w_j > 0} w_j, \quad S(a) = \frac{\sum_j w_j \cdot \mathbb{1}[\text{satisfied}(a, c_j)]}{S_{\text{max}}}
\end{equation}
where $a$ is the agent's deliverable. Catastrophic failure conditions (negative-weight criteria) trigger immediate score penalties regardless of other criteria satisfaction.

\paragraph{Complexity-Adaptive Criterion Count.}
Following the insight from \citet{li2026ares} that rubric granularity should match task complexity, we calibrate criterion count to task difficulty:
\begin{equation}
    |C_{\text{task}}| \approx 10 \cdot \text{difficulty} + \epsilon
\end{equation}
where difficulty $\in \{1, 2, 3, 4, 5\}$ and $\epsilon$ accounts for domain-specific variation. In practice, our rubrics range from 38 criteria (difficulty 2) to 49 criteria (difficulty 5), with a median of 43 criteria.

\subsection{Stage 4: Validation}
\label{sec:stage4}

Generated rubrics undergo a multi-gate validation process combining automated quality assurance with optional expert review:

\paragraph{Gate 1: Automated Quality Assurance.}
All rubrics must pass automated validation checks:
\begin{itemize}
    \item \textbf{Prompt-rubric coverage}: Every explicit requirement in the task prompt has at least one corresponding criterion
    \item \textbf{Deliverable-rubric alignment}: Each criterion is verifiable against the reference deliverable
    \item \textbf{Principle compliance}: Automated checks for evaluability (each criterion has a determinate scoring rule), unambiguity (concrete indicators present), and orthogonality (low inter-criterion semantic overlap)
    \item \textbf{Weight distribution}: Positive weights follow expected distribution (approximately 50\% weight-1, 30\% weight-2, 20\% weight-3+)
    \item \textbf{Discriminative power}: The rubric assigns meaningfully different scores to the reference deliverable versus degraded or model-generated alternatives
\end{itemize}

\paragraph{Gate 2: Discriminative Review.}
The rubric is applied to score deliverables of known different quality (one human expert-authored, one or more model-generated). Evaluators, either domain experts or LLM judges, assess whether the rubric:
\begin{itemize}
    \item Distinguishes the authentic reference from deliberately degraded or materially lower-quality alternatives, and identifies observable quality differences
    \item Identifies specific quality dimensions where alternatives fall short
    \item Does not award points for superficial compliance without substantive content
\end{itemize}
Gate~2 is an internal consistency check on the task-specific rubric layer: it verifies that criteria detect genuine quality gaps, not that any particular author must win. The headline human--agent comparison instead uses the gold-held-out role-level (typical) rubric described below, which does not presuppose that the human deliverable ranks above strong agents. In our experiments, we employ LLM-as-judge \citep{zheng2023judging} for scalable automated discriminative validation.

\paragraph{Gate 3: Cross-Evaluator Consistency.}
Two independent evaluators (human experts or distinct LLM judges with different prompting strategies) score the same deliverable using the rubric. We require:
\begin{equation}
    \text{Cohen's } \kappa \geq 0.75
\end{equation}
following the classical inter-rater agreement coefficient of \citet{cohen1960kappa}. Criteria with low agreement are revised for clarity or removed. This gate ensures that the rubric's formulation is sufficiently unambiguous for reliable automated scoring.

\subsection{From Role-Level to Task-Level Rubrics}

A key advantage of RGRC is the separation between \textit{role-level competency standards} (extracted once per role from the deliverable corpus) and \textit{task-level rubrics} (instantiated per evaluation task). This enables:

\paragraph{Rubric Transfer.}
When creating a new task for a role with an existing competency profile, approximately 60--70\% of criteria can be directly inherited from the role-level standard, with only task-specific criteria (particular data requirements, scenario constraints) requiring de novo generation:
\begin{equation}
    \text{Rubric}_{\text{task}} = \text{Rubric}_{\text{role}} \cup \text{Rubric}_{\text{task-specific}}
\end{equation}

%\begin{figure}[t]
%\centering
%\includegraphics[width=0.88\columnwidth]{Figures/fig3_task_instance_redesign.pdf}
%\caption{Anatomy of a FinProBench task instance for the sell-side equity
%research analyst role (an in-depth report on Meituan-W). \textbf{Top:} the task
%\emph{prompt}, specifying role, scenario, five analytical modules, data and
%compliance constraints, deliverable format, and catastrophic-failure
%conditions. \textbf{Middle:} the \emph{human gold deliverable}, a real
%28-page brokerage research report. \textbf{Bottom:} the RGRC-derived
%\emph{evaluation rubric}, 43 positive criteria (63 points) plus 3 penalty
%conditions, against which an agent's deliverable is scored.}
%\label{fig:example-task}
%\end{figure}

\subsection{Comparison with Existing Methods}
Direct, contrastive, and iterative methods derive criteria from prompts or model outputs \citep{liu2025openrubrics,xie2025autorubric,shen2026rrd}; online variants retain the same source limitation \citep{xu2026sibylsense,li2026ares}. GDPval uses expert-authored task-specific rubrics \citep{patwardhan2025gdpval}, which provide strong targets but no role-level reuse mechanism. RGRC instead combines authentic-deliverable grounding, implicit-standard recovery, and role-to-task reuse with low recurring expert effort.

\section{The FinProBench Benchmark}

We instantiate RGRC as \textbf{FinProBench}, a benchmark for evaluating AI agents on professional financial deliverables. The China-sourced release is the first test bed for a jurisdiction-agnostic, role-grounded construction pipeline.

\subsection{Corpus Construction}

The present instantiation comprises 1,723 authentic professional documents collected from public authoritative sources in China's financial industry. The corpus spans 8 sub-industries aligned with the GB/T 4754-2017 national standard: banking (monetary and financial services), securities (capital markets), insurance, asset management, wealth management, fintech, compliance, and data engineering. Within these industries, we cover 57 distinct professional roles spanning 161 deliverable types, including equity research analysts, credit rating analysts, fund managers, compliance officers, financial engineers, and insurance actuaries.

Each document passes three quality gates: format validation (well-formed PDF, $\geq$800 extractable characters), cryptographic deduplication (MD5), and authenticity verification per the Deliverable Collection stage (Stage~1). The final corpus is duplicate-free.

\subsection{Task Construction}

Each FinProBench evaluation task is a triple $(p, d^*, R)$ consisting of:
\begin{itemize}
    \item A \textbf{task prompt} $p$ specifying the role, scenario, analytical requirements, data constraints, deliverable format, and catastrophic-failure conditions
    \item A \textbf{human gold deliverable} $d^*$ produced by a practitioner in the target role
    \item An \textbf{RGRC-derived rubric} $R = \{(c_j, w_j, t_j)\}_{j=1}^{|R|}$ with criteria, weights, and category tags
\end{itemize}

We construct 20 complete tasks spanning 20 roles across 7 sub-industries. Task complexity ranges from difficulty level 2 (a fund quarterly report, 38 criteria) to level 5 (the most demanding research tasks, 49 criteria), with a median of 43 criteria per task.

\subsection{Rubric Statistics}

Across the 20 tasks, final task rubrics contain 38--49 criteria (median 43), including a median of 3 penalty conditions. The role-level criteria reuse rate averages 64.7\%, with 35.3\% of criteria requiring task-specific generation.

\section{Experiments}

We evaluate FinProBench with GPT, DeepSeek, Qwen, and Gemini judges, using identical role-level rubrics and rollout protocols for all systems. We organize the analysis around reliability, validity, and the human--agent comparison. Each criterion is scored by a unified panel of four heterogeneous frontier LLM judges from distinct families: GPT, DeepSeek, Qwen, and Gemini. The full rubric is batched into one prompt, and each judge returns a binary decision for every criterion. We repeat scoring over three rollouts to quantify agreement within and across judges. A dedicated self-preference analysis (drop-one-provider re-scoring) finds no material own-model bias: within-family effects stay inside the judge-variance band and are an order of magnitude smaller than differences in judge strictness. We therefore score every system with the same four judges and report same-provider exclusion only as a robustness check.

\textbf{Rubric quality.} All 20 rubrics pass Gate 1 automated quality assurance: prompt--rubric coverage is 100\% (every explicit requirement maps to $\geq 1$ criterion, enforced by construction), criteria carry determinate, objective scoring rules (anchorless subjective items are flagged for revision), and inter-criterion semantic overlap is low, indicating minimal redundancy.

\paragraph{Metrics.} Each criterion receives one binary judgment per judge: a positive criterion counts as met when its awarded score reaches at least half its weight, and a negative criterion fires when triggered. A deliverable's task score is the weighted satisfaction ratio $S(a)$ of Eq.~(5) rescaled to $[0,100]$, averaged over the three rollouts and the four judges; the \emph{Mean Score} is the macro-average of these per-task scores across the 20 tasks. \emph{Tasks Won} counts, per task, the system with the highest mean score, assigning exact ties to the human anchor. Reported confidence intervals are Student-$t$ intervals $\bar{x}\pm t_{0.975,\,n-1}\,s/\sqrt{n}$ with $n=20$. The per-dimension \emph{positive-hit rate} (Fig.~\ref{fig:dimensions}) is the weight-pooled attainment $100\cdot\sum_j \mathrm{awarded}_j/\sum_j \mathrm{max}_j$ over the positive-weight criteria mapped to that dimension. Inter-judge reliability is Fleiss' $\kappa$ over the binarized criterion labels, taking each judge's majority label across rollouts and treating every (task, system, criterion) triple as a subject.

\subsection{Professional-Standard Coverage}
Before coverage analysis, we classified all 57 occupations using only deliverable-genre characteristics, independently of RGRC--baseline outcomes. Prior-rich conventional roles (30) use genres widely available on the public internet, research reports, product documentation, marketing materials, and technical documentation, so general-model priors are strong. Prior-sparse role-specialized roles (27) rely on regulatory or statutory formats, formal methodologies, or institution-internal professional genres, for which public priors are thin. Held-out coverage is evaluated across all 30 conventional and 27 role-specialized roles. Prompt-only sees only the task prompt, whereas RGRC is grounded in authentic professional work products. Evaluation standards are independently extracted from held-out documents of the same type, with all held-out documents audited against the RGRC source pool.

\begin{table}[t]
\centering
\caption{Budget-matched held-out professional-standard coverage. All 57 occupations were classified before analysis into 30 prior-rich conventional and 27 prior-sparse role-specialized roles. Coverage is aggregated over all roles in each preclassified group.}
\label{tab:rubric_baselines}
\scriptsize
\small\setlength{\tabcolsep}{2.4pt}
\begin{tabular}{@{}llccc@{}}
\toprule
\textbf{Regime} & \textbf{Method} & \textbf{\#Crit.} & \textbf{Cov.} & \textbf{95\% CI} \\
\midrule
Conventional (30) & Prompt@78 & 79.2 & 89.2\% & [76.7, 100] \\
 & RGRC & 78.2 & 90.7\% & [79.4, 100] \\ \midrule Role-spec. (27) & Prompt@78 & 79.2 & 78.0\% & [66.8, 89.1] \\
 & \textbf{RGRC} & 79.3 & \textbf{99.1\%} & [97.8, 100] \\
\bottomrule
\end{tabular}
\end{table}

Controlling criterion budget reveals a regime-dependent result. Across the 30 conventional roles, Prompt@78 nearly matches RGRC (89.2\% versus 90.7\%; a 1.5-point gap), so we do not claim universal superiority. Across the 27 role-specialized roles, however, RGRC reaches 99.1\% held-out coverage versus 78.0\% for Prompt@78, a 21.2-point advantage. This interaction is the central finding: prompt engineering can approximate professional rubrics when task conventions are already encoded in general-model priors, whereas RGRC is essential when evaluation depends on tacit, role-specific standards that prompts alone cannot recover. Its value lies not in universal rubric expansion but in supplying external professional evidence beyond model priors.

\subsection{Judge Reliability}

A central concern is that criteria encoding \emph{implicit professional standards}, the distinctive value of deliverable-grounded rubrics, may be scored less consistently than explicit requirements. Across the 20 tasks, however, independent judges reach a high mean criterion-level agreement: Fleiss' $\kappa = 0.76$ on binarized criterion judgments (raw inter-judge agreement $89.5\%$), meeting our $\kappa \geq 0.75$ consistency gate, indicating that RGRC criteria are formulated unambiguously enough for reliable automated scoring.

\subsection{Discriminative Validity and the Human--Agent Gap}

Every FinProBench task includes an \emph{authentic human-authored deliverable}, which we use as an embedded validity anchor without separate expert annotation. We distinguish two rubric views. The \emph{task rubric} is instantiated from the specific gold artifact, making the human deliverable its effective reference answer. A human win under this rubric is therefore close to tautological. We use the task rubric only as an internal consistency check and \emph{not} as evidence of discriminative power. The \emph{gold-held-out role-level rubric} (\emph{typical rubric} for short) is distilled from an independent corpus of the same document type rather than from the specific gold artifact, so no system is privileged by construction. All headline results below are reported under the typical rubric. We score the human gold alongside three general-purpose agents (Qoder, Codex, and Claude Code), each run with the same task prompt and output requirements. 

\begin{table}[t]
\centering
\caption{Evaluation under the \emph{gold-held-out} role-level rubric, averaged over 20 tasks with three rollouts and a unified four-judge panel (GPT, DeepSeek, Qwen, Gemini). All three agents follow comparable protocols. \emph{Won} counts per-task highest mean scores, assigning exact ties to the human anchor.}
\label{tab:agent_results}
\small
\begin{tabular}{@{}lccc@{}}
\toprule
\textbf{System} & \textbf{Mean} & \textbf{95\% CI} & \textbf{Won} \\
\midrule
Human deliverable & \textbf{73.7} & [68.5, 78.9] & \textbf{8} \\
\midrule
Codex & 70.3 & [64.4, 76.2] & 3 \\
Qoder & 70.2 & [65.6, 74.7] & 2 \\
Claude Code & 69.6 & [62.7, 76.6] & 7 \\
\bottomrule
\end{tabular}
\end{table}

\begin{figure}[t]
\centering
\includegraphics[width=0.86\columnwidth]{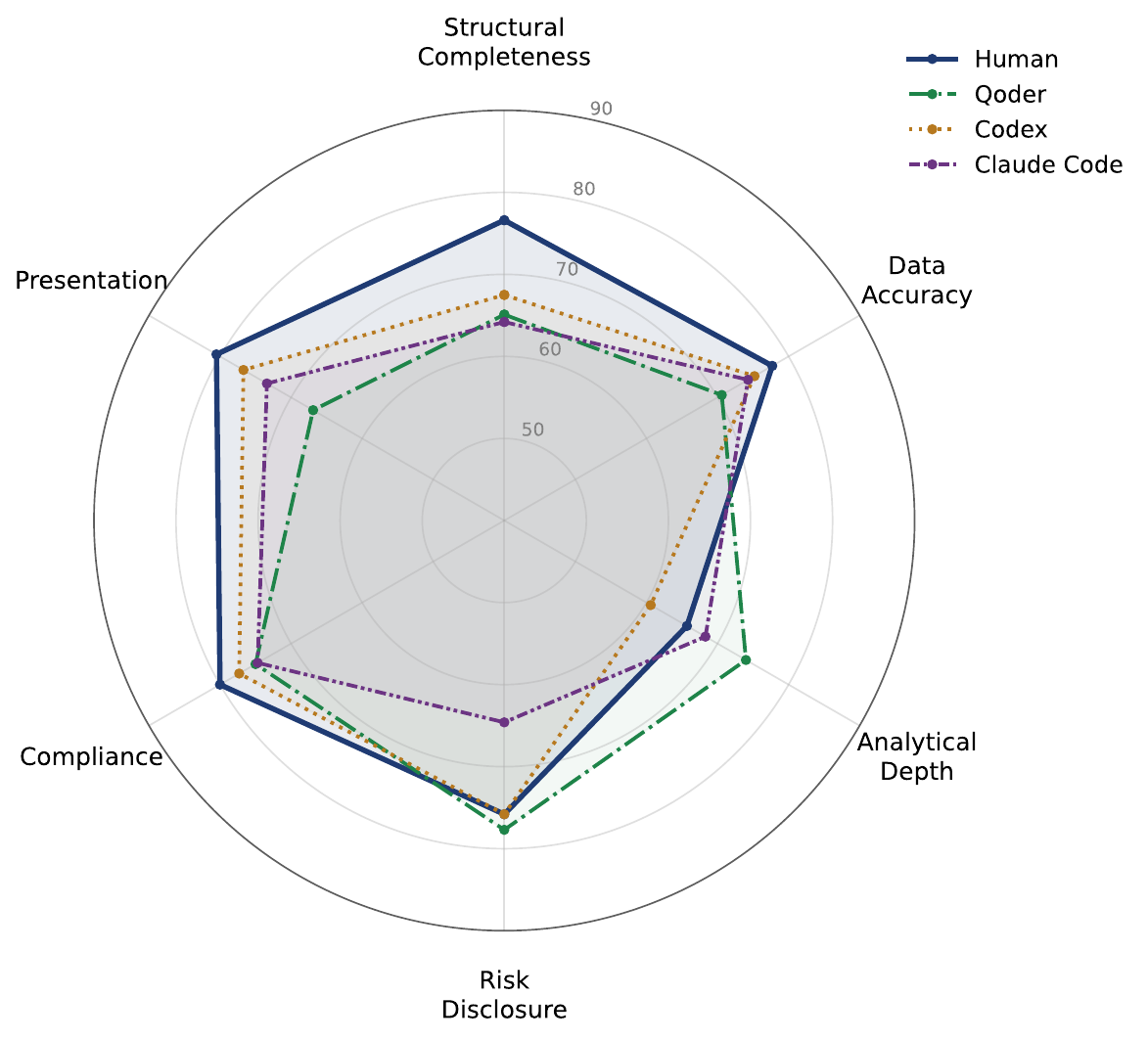}
\caption{Per-dimension positive-hit rate (\%) under the gold-held-out role-level rubric. Humans lead structural completeness, data accuracy, compliance, and presentation; Qoder leads analytical depth and risk disclosure. No system dominates every dimension.}
\label{fig:dimensions}
\end{figure}

Under the gold-held-out role-level rubric, human deliverables lead on average (73.7 versus 70.3 for Codex, 70.2 for Qoder, and 69.6 for Claude Code). The four 95\% confidence intervals overlap at $n=20$, so these are descriptive rankings rather than significance claims. Per-task wins are distributed across all four systems: Human wins eight, Claude Code seven, Codex three, and Qoder two. Claude Code combines the lowest agent mean with the widest interval, indicating high task-level variance. Dimension profiles are complementary (Fig.~\ref{fig:dimensions}). Humans lead structural completeness, data accuracy, compliance, and presentation, whereas Qoder leads analytical depth and risk disclosure. Codex is closest to the human profile on data, compliance, and presentation. Claude Code combines comparatively strong analytical depth with high cross-task variance. Under same-source exclusion (the thinner DeepSeek--Gemini intersection), humans remain first (74.0), so the ranking is not a panel artifact. These profiles clarify why the task-level averages are close despite distinct system behaviors. Human performance reflects broad consistency across document structure, evidence handling, compliance, and presentation, whereas agent advantages are concentrated in fewer dimensions. Reporting both aggregate scores and dimension profiles therefore distinguishes complete-work-product quality from isolated analytical strengths and avoids reducing the comparison to a single leaderboard value.

\section{Limitations}
Our validation combines held-out professional deliverables with a heterogeneous LLM judge panel but lacks independent practitioner annotation. The results therefore establish automated coverage, discriminative validity, and evaluator consistency rather than practitioner consensus. Because the evidence corpus consists of public professional deliverables, it may overrepresent externally documented roles and conventions. Cross-jurisdictional evaluation should test the stability of role-grounded criteria across languages, regulatory regimes, and institutions.
\paragraph{Conclusion.} We introduced FinProBench and RGRC, deriving criteria from authentic deliverables and reusing role-level standards across tasks. Held-out role-level rubrics distinguish human and agent work, while controlled baselines show that RGRC's main gain is recovering standards beyond general-model priors. Human deliverables lead on average, while Codex, Qoder, and Claude Code show distinct task-level and dimension-level strengths; no individual system dominates the complete-work-product evaluation. Although instantiated with China-sourced financial deliverables, RGRC is jurisdiction-agnostic: extension requires a new role-aligned evidence corpus and regulatory criteria, not a new rubric-construction methodology.

\paragraph{Use of AI Systems.} Generative-AI systems served as rubric generators and judges; they also assisted code generation, language polishing, and figure preparation. The authors designed the study, verified sources and results, reviewed all AI-assisted material, and take full responsibility; no AI system is an author or cited source.
\clearpage
\bibliography{aaai2027}

\end{document}